\documentclass[11pt, conference]{IEEEtran}
\IEEEoverridecommandlockouts
\usepackage{graphicx}
\usepackage{cite}
\usepackage{amsmath,amssymb,amsfonts}
\usepackage{algorithmic}
\usepackage{graphicx}
\usepackage{textcomp}
\usepackage{xcolor}
\usepackage{multirow}
\usepackage{hyperref}
\usepackage{caption}
\usepackage{booktabs}
\def\BibTeX{{\rm B\kern-.05em{\sc i\kern-.025em b}\kern-.08em
    T\kern-.1667em\lower.7ex\hbox{E}\kern-.125emX}}
\begin{document}

\title{Integrating Frequency-Domain Representations with Low-Rank Adaptation in Vision-Language Models}

\author{%
  {Md Azim Khan\textsuperscript{1,2}}, 
  {Aryya Gangopadhyay\textsuperscript{1,2}}, {Jianwu Wang\textsuperscript{1,2}}, {Robert F. Erbacher\textsuperscript{3}} \\
  \textit{\textsuperscript{1}University of Maryland Baltimore County (UMBC), USA}\\
  \textit{\textsuperscript{2}Center for Real-time Distributed Sensing and Autonomy (CARDS), UMBC, USA}\\
   \textit{\textsuperscript{3}DEVCOM Army Research Laboratory, Adelphi, Maryland, USA}\\
  \textit{\textsuperscript{1,2}(azimkhan22, gangopad, jianwu)@umbc.edu}, robert.f.erbacher.civ@army.mil}

\maketitle

\begin{abstract}
Situational awareness applications rely heavily on real-time processing of visual and textual data to provide actionable insights. Vision language models (VLMs) have become essential tools for interpreting complex environments by connecting visual inputs with natural language descriptions. However, these models often face computational challenges, especially when required to perform efficiently in real environments. This research presents a novel vision language model (VLM) framework that leverages frequency domain transformations and low-rank adaptation (LoRA) to enhance feature extraction, scalability, and efficiency. Unlike traditional VLMs, which rely solely on spatial-domain representations, our approach incorporates Discrete Fourier Transform (DFT)-based low-rank features while retaining pretrained spatial weights, enabling robust performance in noisy or low-visibility scenarios. We evaluated the proposed model on caption generation and Visual Question Answering (VQA) tasks using benchmark datasets with varying levels of Gaussian noise. Quantitative results demonstrate that our model achieves evaluation metrics comparable to state-of-the-art VLMs, such as CLIP ViT-L/14 and SigLIP. Qualitative analysis further reveals that our model provides more detailed and contextually relevant responses, particularly for real-world images captured by a RealSense camera mounted on an Unmanned Ground Vehicle (UGV).

\end{abstract}


\section{Introduction}
\label{intro}

Understanding the surrounding environment is crucial for a robot to determine its next steps, such as whether to proceed or halt. Robots can analyze their surroundings to gain contextual understanding and make informed decisions. This capability helps bridge the gap between human and machine interactions. It also improves situational awareness in complex and dynamic environments. Vision language models (VLM) \cite{zhang2024vision} have emerged as a growing area of research, combining the strengths of computer vision and natural language processing.

However, traditional Vision-Language Models (VLMs) often face challenges related to computational efficiency and performance, particularly in adverse conditions such as low visibility or noisy data. The development of parameter-efficient vision language models has gained significant attention, addressing the need for scalable and adaptable multimodal learning in real-world applications. Low-Rank Adaptation (LoRA) \cite{hu2021lora} is one such method that enables efficient fine-tuning by inserting low-rank trainable matrices into pre-trained model weights. The core of low-rank adaptation is Singular Value Decomposition (SVD)\cite{kalman1996singularly}, which decomposes a matrix into three components to optimize its structure. In paper \cite{zhang2023adalora}, the authors employ low-rank techniques in language models to optimize weight matrices by scoring eigenvectors and singular values. Another study ranks spaces according to orthogonality, allowing models to adapt to new tasks while minimizing the risk of catastrophic forgetting \cite{goldfarb2023analysis}. However, these approaches typically apply low-rank methods to the weight matrix space alone, which limits their ability to capture complementary features, especially in noisy or high-dimensional data scenarios.

To address these challenges, we propose the integration of frequency domain transformations with low-rank adaptation in vision language models (VLM) to achieve robust and efficient multimodal learning.  The frequency transformation provides additional information about the data by uncovering additional information that enhances understanding. Frequency-domain representations excel at capturing global patterns, while low-rank adaptation ensures scalability and parameter efficiency. In summary, our main contributions are as follows.

\begin{itemize}
\item Our approach leverages DFT-based low-rank frequency features while preserving essential spatial information and retaining the pre-trained model weights.
\item We validated our approach on benchmark datasets, demonstrating its robustness and efficiency in processing visual inputs and generating descriptive outputs under varying levels of Gaussian noise and feature matrix ranks.

\item By combining frequency-based visual features with the Phi-2 language model\cite{li2023textbooks}, our model generates contextual scene descriptions and insightful answers for VQA tasks, enhancing scene understanding for real-time images.

\end{itemize}

\section{Related Work}
\label{rw}
Vision-Language Models (VLMs) have seen significant advancements through techniques aimed at improving efficiency and scalability. Recent works like CoCa \cite{yu2022coca} and Beit-V3\cite{wang2023image} have introduced feature refinement strategies that integrate multimodal features at various levels of abstraction, enhancing contextual understanding and adaptability to tasks. An adapter\cite{wang2020k}, lightweight modular architectures, have further optimized task-specific fine-tuning by adding small, trainable modules to transformer layers. Similarly, the author's approaches in \cite{li2021align}, focus on aligning vision and language modalities with minimal parameter changes, achieving robust performance in tasks such as image captioning \cite{hossain2019comprehensive} and Visual Question Answering (VQA) \cite{kafle2017visual}. Research based on sparse methods \cite{zaken2021bitfit} optimizes the bias parameter to reduce cost and improve the performance of the model during specific tasks but faces problems in dealing with real data.
The paper \cite{kudugunta2021beyond} demonstrates that the performance of the model can be improved using a mixture of experts system, which is a sparse method that scales the capacity of the model while incurring lower training costs compared to dense variants. Furthermore, energy-efficient approaches such as MiniGPT \cite{zhu2023minigpt} and techniques such as quantization and pruning \cite{shen2023pruning} have optimized resource utilization for deployment on robotic platforms and mobile devices.

Frequency domain transformations offer another promising avenue for improving model robustness \cite{frank1994frequency}. These transformations convert time domain signals to frequency domain \cite{peters1998discrete} and have been successfully applied in tasks such as image compression, noise removal, texture analysis and noise suppression\cite{vilar2001noise}, \cite{tuceryan1993texture}. These techniques capture global and structural features, making them particularly effective in multimodal contexts. Low-Rank Adaptation (LoRA)\cite{hu2021lora} complements these efforts by introducing low-rank matrices to adapt the pretrained model parameters. The author in paper \cite{sun2017sparse} employed the low-rank approximation to compress image data, preserving essential patterns while reducing dimensionality. Low-rank approximation within convolutional neural networks (CNNs) for image super-resolution applied to achieve detailed outputs critical for applications like medical imaging and surveillance, but struggle with extreme noise \cite{hansen1987truncated}.

The integration of frequency-domain transformations with low-rank adaptation into Vision-Language Models (VLMs) offers new opportunities for robust and efficient multimodal learning. Tasks like VQA and image captioning, which require models to process intricate visual-textual relationships, can benefit significantly from this combination. Using frequency domain transformations to capture global patterns and low-rank adaptation for scalability, our approach enhances feature extraction and task performance, addressing key limitations in traditional methods.

\section{Methodology}
\label{met}
\subsection{Problem Formulation}
In the low-rank approximation, the main concept is Singular Value Decomposition (SVD) which decomposes a matrix \( M \) into three components \( M = U \Sigma V^T \), where \( U \) and \( V \) are orthogonal matrices, and \( \Sigma \) is a diagonal matrix containing the singular values of \( M \) \cite{eckart1936approximation}. These singular values provide insight into the intrinsic structure of the data, with larger singular values corresponding to more significant components. By truncating the smaller singular values and the corresponding vectors, we can approximate \( M \) as \( M \approx L_k R_k^T \), where \( L_k \) and \( R_k \) represent the singular vectors left and right corresponding to the top singular values \( k \). 

\par In the regular fine-tuning process, the pre-trained weights \( W \) are updated using a weight update matrix \( \Delta W \). This results in new weights \( W' = W + \Delta W \). The updated weights \( W' \) are then multiplied by the input \( x \) to produce the output \( h' \):

\begin{equation}
h' = W' \times x = (W + \Delta W) \times x
\label{eq:regular_finetuning}
\end{equation}

In the LoRA\cite{hu2021lora} process, instead of directly updating the weight matrix, the weight update \( \Delta W \) is approximated using two lower dimensional matrices \( A \) and \( B \). The matrix \( A \) has dimensions \( u_1 \times k \) and the matrix \( B \) has dimensions \( k \times u_2 \), where \( k \) represents the low-rank dimension. The new weights \( W' \)  and the output \( h' \) after applying the input \( x \) are calculated as:

\begin{equation}
\Delta W = A \times B
\label{eq:lora_deltaW}
\end{equation}

\begin{equation}
W' = W + \Delta W = W + A \times B
\label{eq:lora_new_weights}
\end{equation}

\begin{equation}
h' = W' \times x = (W + A \times B) \times x = W \times x + A \times (B \times x)
\label{eq:lora_output}
\end{equation}

This formulation allows the model to adapt to new tasks by adjusting only a small number of parameters in the low-rank matrices \( A \) and \( B \), rather than the entire weight matrix \( W \).

\subsection{Proposed Architecture}

\par
\begin{figure}[!ht]
\centering
\includegraphics[width=\linewidth]{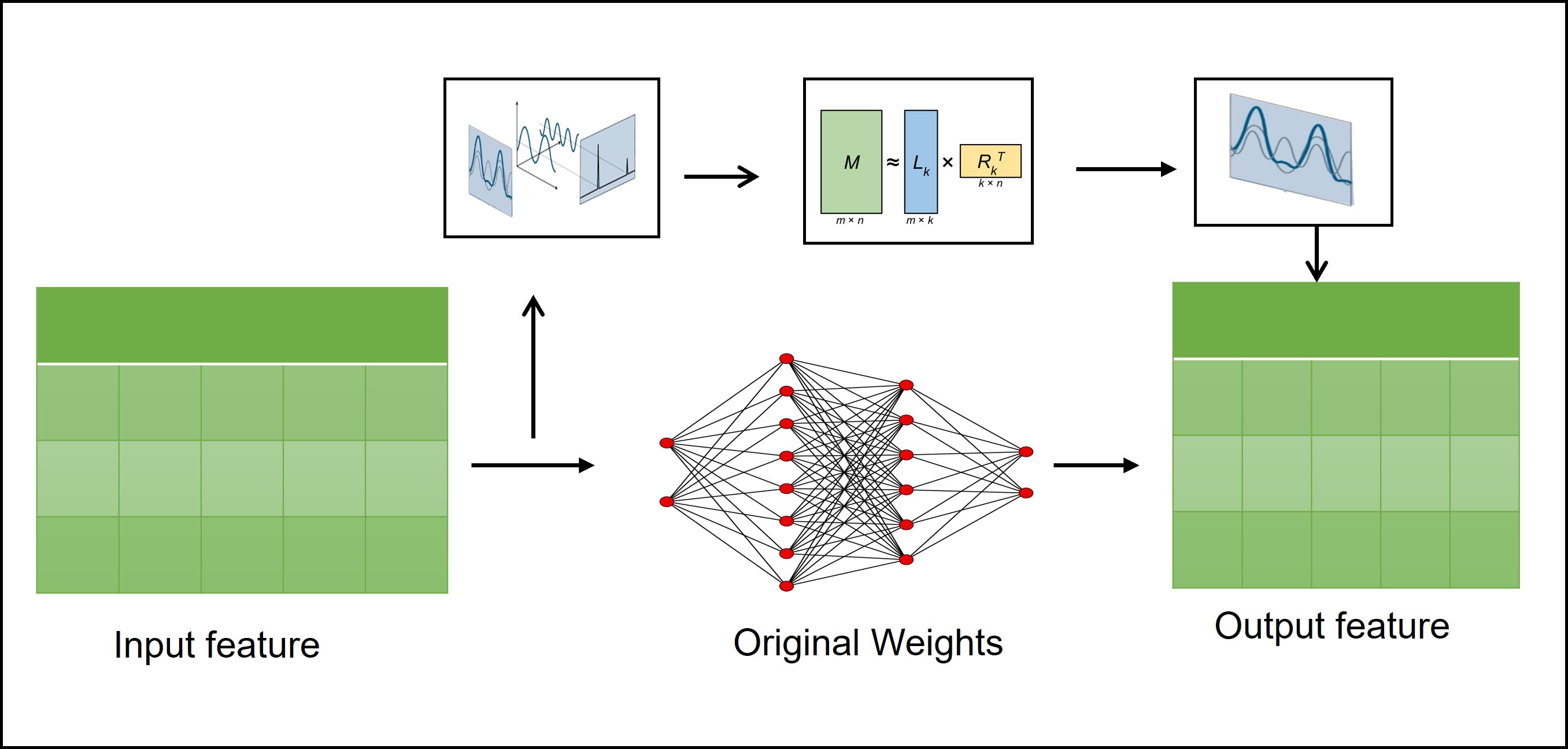}
\caption{Proposed Architecture}
\label{loradft}
\end{figure}

 In our proposed architecture, as shown in Figure \ref{loradft}, we follow a process similar to LoRA, where instead of directly updating the weight matrix \(W\), we add a parallel branch that takes the input features and converts them into the frequency domain, as described in Equation 5. The main difference between Equations 4 and 5 is that both adjust a small number of parameters in low-rank matrices A and B, but one operates in the spatial domain, while the other operates in the frequency domain.

Given an input feature matrix \( x\), the output after applying the low-rank adaptation, denoted h', is calculated as follows:

\begin{equation}
    h' = W  \times x + \mathcal{F}^{-1}(\alpha B A \mathcal{F}(x))
    \label{eq:lora_equation}
\end{equation}

Here, \( \mathcal{F}(\cdot) \) and \( \mathcal{F}^{-1}(\cdot) \) represent the discrete Fourier transforms and the inverse frequency transforms, respectively. And \( \alpha \) is a hyperparameter that controls the frequency and inverse frequency relation and provides flexibility in controlling the spectral emphasis for diverse datasets
without modifying BA.
To reduce noise in the image as well as minimize information loss during the projection into the low-rank subspace, the inputs are first transformed into the frequency domain using the Discrete Fourier Transform (DFT). In the frequency domain, a low-rank approximation is applied to the transformed input features. Performing low-rank adaptation in the frequency domain provides noise robustness, and the robustness of our approach is demonstrated in Section IV by manually adding Gaussian noise and real-world images captured using a Real-Sense camera. After the low-rank approximation, the frequency domain representation is transformed back into the spatial domain using the inverse DFT (IDFT). The idea is to retain only the most significant frequencies, reducing the dimensionality while preserving essential information.

\section{Experiments}
\label{exp}

We began by testing our proposed architecture on the vision component of a Vision-Language Model (VLM). For this, we used vision modules such as CLIP ViT-L / 14 \cite{radford2021learning} and SigLIP \cite{zhai2023sigmoid}, modifying them to integrate our DFT + LoRA approach. Our experiments included three setups: normal fine-tuning, applying LoRA, and applying DFT + LoRA. To evaluate robustness, we introduced Gaussian noise during training and testing, observing how each set-up performed under varying noise conditions. These experiments were conducted separately for the CLIP and SigLIP vision modules.

\begin{figure}[!ht]
\centering
\includegraphics[width=\linewidth]{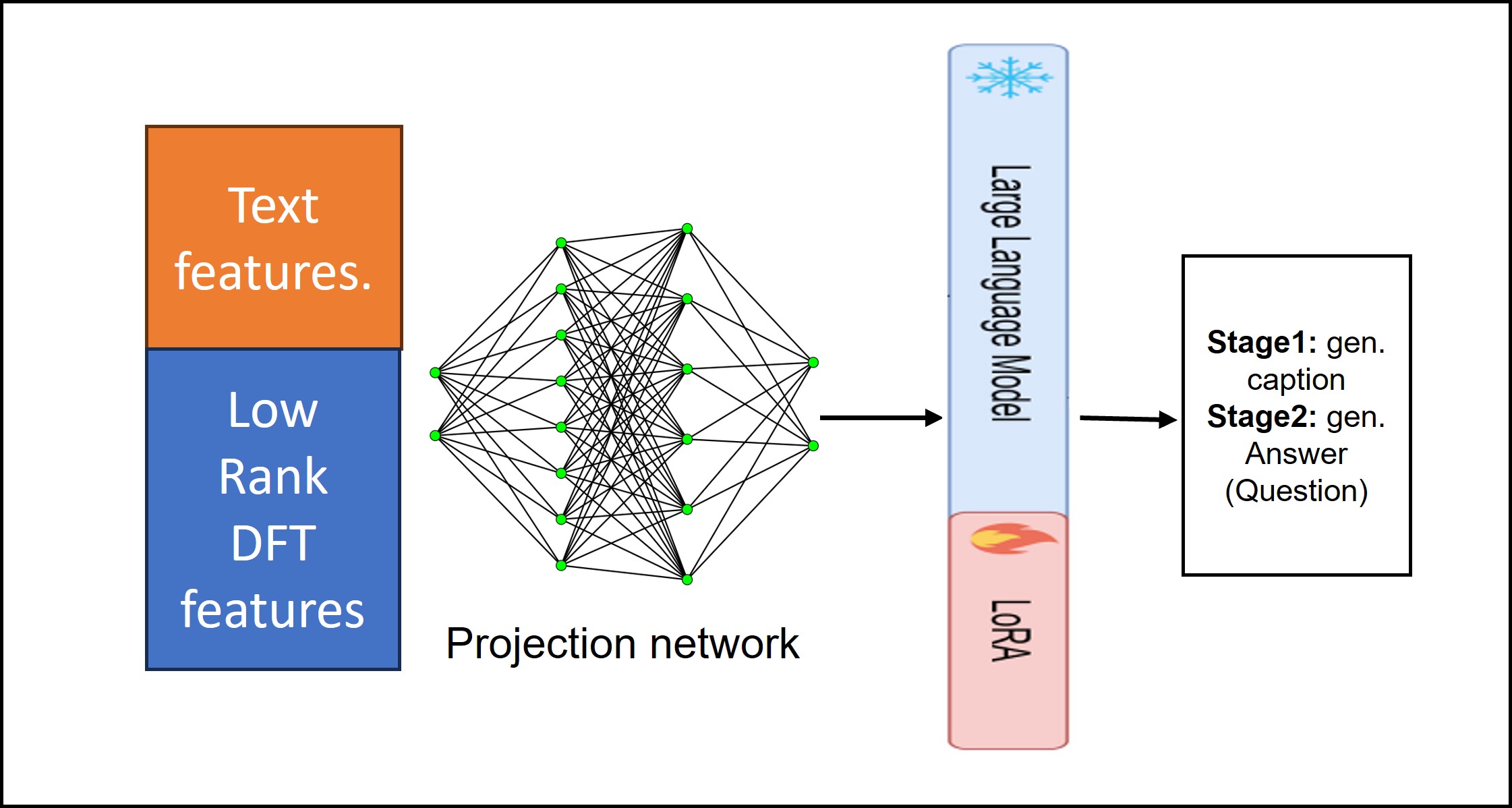}
\caption{Framework of the proposed Vision-Language Model (VLM) integrating low-rank DFT features with text embeddings via a projection network. The model employs a two-stage training process: Stage 1 for caption generation and Stage 2 for Visual Question Answering (VQA). The LoRA adaptation is applied to enhance task-specific performance}
\label{experiment}
\end{figure}

Building on these experiments, we constructed a complete VLM model, as illustrated in Figure \ref{experiment}, incorporating the features DFT + LoRA in the framework. The training process followed a two-stage strategy inspired by Llava1.5 \cite{liu2024visual}. In the first stage, the model was trained on the COCO 2017 dataset \cite{lin2014microsoft}, which includes 118,287 images and 591,753 captions, to generate descriptive captions. In the second stage, the Phi2 language model was fine-tuned by using low-rank adaptation to enhance its capabilities for Visual Question Answering (VQA). This involved fine-tuning both the projection network and Phi2 in data sets such as QAv2\cite{antol2015vqa}, GQA\cite{hudson2019gqa}, and TextVQA\cite{singh2019towards}. Model performance was evaluated using standard metrics (BLEU, ROUGE, METEOR, CIDEr) for caption generation and VQA accuracy to assess understanding and reasoning capabilities. Our training objective for caption generation was to minimize the negative log-likelihood of the sequence. For Visual Question Answering (VQA), we used cross-entropy loss, treating it as a classification task where the model predicts the correct answer from a set of options.

The implementation is built using PyTorch, and all experiments were performed on NVIDIA RTX A6000 GPUs. AdamW is used as the optimizer, in combination with the OneCycleLR learning rate scheduler, with a maximum learning rate set to $1 \times 10^{-4}$, following a cosine annealing strategy. The size of the image patch is configured as 378, with the input dimensions set to $(378 \times 378)$. The Phi-2 model is configured to balance efficiency and representational power, ensuring optimal performance for caption generation and Visual Question Answering (VQA) tasks. Key parameters include a vocabulary size of 51,200, a hidden size of 2048, and an intermediate size of 8192, with 24 hidden layers and 32 attention heads.

\section{Results and Discussion}

We tested two vision modules, CLIP ViT-L/14 and SigLIP, under varying noise conditions, which are the backbone of the Llava model. Tables \ref{siglip}, \ref{clip} present a comparative performance of three configurations: normal fine-tuning, applying LoRA, and applying DFT + LoRA. The results indicate that while normal fine-tuning provides a baseline for performance, the addition of LoRA introduces a notable improvement across all evaluation metrics, particularly under no-noise and low-noise scenarios. However, the integration of DFT + LoRA further enhances the model's robustness, particularly in high-noise conditions, as observed from higher BLEU and CIDEr scores. This improvement highlights the importance of combining frequency-domain features with low-rank adaptation for better feature representation.

For both SigLIP and CLIP ViT-L/14, as shown in Tables I and II, the DFT + LoRA approach achieved a 6\% increase in BLEU-4 and a 4\% improvement in CIDEr scores on average, indicating better caption generation capabilities. Metrics such as ROUGE and METEOR also showed notable gains, with an average improvement 3\% in ROUGE and 2\% in METEOR compared to the baseline. In VQA tasks, the DFT + LoRA configuration achieved up to a 4\% higher accuracy than the baseline, with a standard deviation of ±0.5\% under no noise conditions and ±0.9\% under noisy environments.

 \begin{figure}[!ht]
\centering
\includegraphics[width=\linewidth]{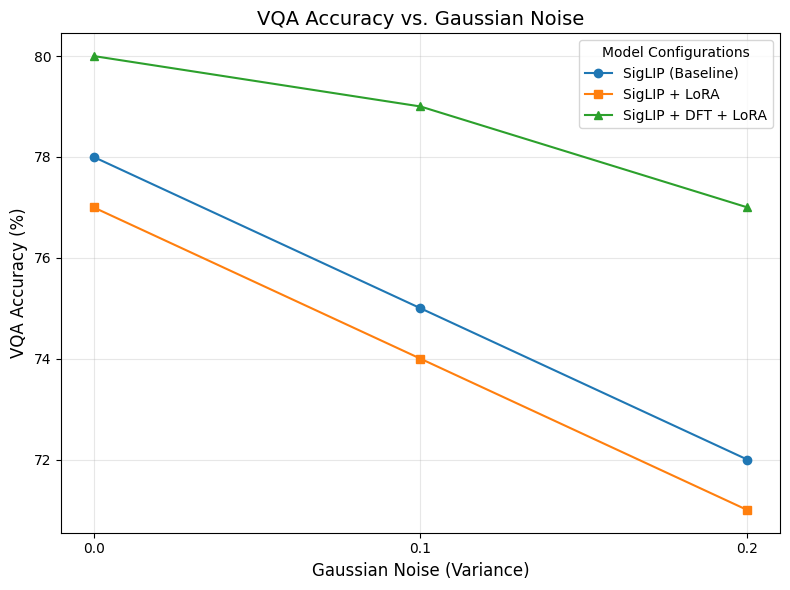}
\caption{Accuracy of VQA on variation of noise}
\label{noise}
\end{figure}

\begin{table*}[t]
\caption{Comparison of Siglip model performance under different noise levels and configurations.}
\centering
\renewcommand{\arraystretch}{1.2}

\setlength{\tabcolsep}{4pt} 
\resizebox{\textwidth}{!}{%
\begin{tabular}{|l|c|c|c|c|c|c|c|c|c|}
\hline
\textbf{Model Configuration} & \textbf{Noise Level} & \textbf{BLEU-1} & \textbf{BLEU-2} & \textbf{BLEU-3} & \textbf{BLEU-4} & \textbf{CIDEr} & \textbf{ROUGE} & \textbf{METEOR} & \textbf{VQA Accuracy (Std)} \\ \hline
SigLIP (Baseline)            & 0 (No Noise)         & 0.72            & 0.55            & 0.42            & 0.31            & 1.15           & 0.58           & 0.27            & 78\% (±0.5)               \\ \cline{2-10}
                             & Variance = 0.1       & 0.68            & 0.50            & 0.39            & 0.28            & 1.10           & 0.53           & 0.25            & 75\% (±1.5)               \\ \cline{2-10}
                             & Variance = 0.2       & 0.65            & 0.48            & 0.37            & 0.26            & 1.05           & 0.50           & 0.23            & 72\% (±2.5)               \\ \hline
SigLIP + LoRA                & 0 (No Noise)         & 0.71            & 0.54            & 0.41            & 0.30            & 1.12           & 0.56           & 0.26            & 77\% (±0.5)                \\ \cline{2-10}
                             & Variance = 0.1       & 0.69            & 0.52            & 0.40            & 0.29            & 1.08           & 0.54           & 0.25            & 74\% (±1.2)               \\ \cline{2-10}
                             & Variance = 0.2       & 0.66            & 0.49            & 0.38            & 0.27            & 1.03           & 0.52           & 0.23            & 71\% (±2.1)               \\ \hline
SigLIP + DFT + LoRA          & 0 (No Noise)         & 0.73            & 0.56            & 0.44            & 0.33            & 1.20           & 0.60           & 0.29            & 80\% (±0.5)               \\ \cline{2-10}
                             & Variance = 0.1       & 0.72            & 0.55            & 0.43            & 0.32            & 1.18           & 0.59           & 0.28            & 79\% (±0.7)               \\ \cline{2-10}
                             & Variance = 0.2       & 0.70            & 0.53            & 0.41            & 0.31            & 1.15           & 0.57           & 0.26            & 77\% (±0.9)               \\ \hline
\end{tabular}
}
\label{siglip}
\end{table*}

\begin{table*}[t]
\caption{Comparison of CLIP ViT-L/14 model performance under different noise levels and configurations.}
\centering
\renewcommand{\arraystretch}{1.2} 

\setlength{\tabcolsep}{4pt} 
\resizebox{\textwidth}{!}{%
\begin{tabular}{|l|c|c|c|c|c|c|c|c|c|}
\hline
\textbf{Model Configuration} & \textbf{Noise Level} & \textbf{BLEU-1} & \textbf{BLEU-2} & \textbf{BLEU-3} & \textbf{BLEU-4} & \textbf{CIDEr} & \textbf{ROUGE} & \textbf{METEOR} & \textbf{VQA Accuracy (Std)} \\ \hline
CLIP ViT-L/14 & 0 (No Noise) & 0.70 & 0.53 & 0.40 & 0.29 & 1.12 & 0.56 & 0.26 & 76\%(±0.5) \\ 
                         & Variance = 0.1 & 0.67 & 0.51 & 0.38 & 0.28 & 1.08 & 0.54 & 0.25 & 74\%(±2.1) \\ 
                         & Variance = 0.2 & 0.65 & 0.49 & 0.36 & 0.27 & 1.05 & 0.52 & 0.24 & 72\%(±2.5) \\ \hline
CLIP ViT-L/14 + LoRA     & 0 (No Noise) & 0.72 & 0.55 & 0.42 & 0.31 & 1.15 & 0.58 & 0.27 & 78\%(±0.5) \\ 
                         & Variance = 0.1 & 0.70 & 0.53 & 0.41 & 0.30 & 1.12 & 0.56 & 0.26 & 76\%(±1.7) \\ 
                         & Variance = 0.2 & 0.68 & 0.51 & 0.39 & 0.29 & 1.10 & 0.55 & 0.25 & 74\%(±2.3) \\ \hline
CLIP ViT-L/14 + DFT + LoRA & 0 (No Noise) & 0.74 & 0.57 & 0.44 & 0.33 & 1.20 & 0.60 & 0.29 & 80\%(±0.5) \\ 
                         & Variance = 0.1 & 0.73 & 0.56 & 0.43 & 0.32 & 1.18 & 0.59 & 0.28 & 79\%(±0.8) \\ 
                         & Variance = 0.2 & 0.71 & 0.54 & 0.41 & 0.31 & 1.15 & 0.58 & 0.27 & 77\%(±0.9) \\ \hline
\end{tabular}%
}
\label{clip}
\end{table*}

\begin{figure}[!ht]
\centering
\includegraphics[width=\linewidth]{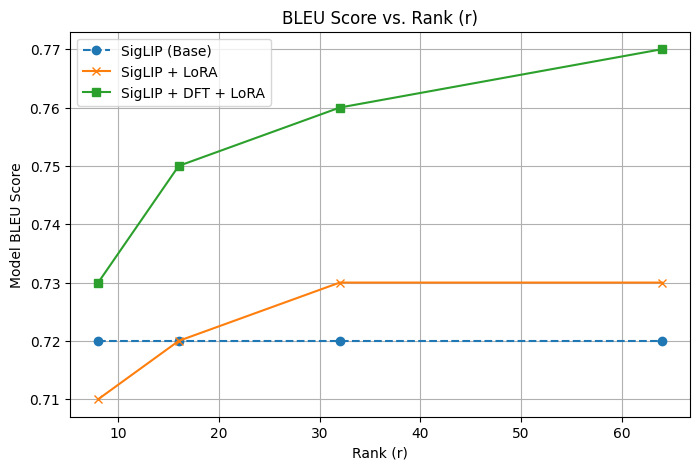}
\caption{Model performance on variation of rank}
\label{blue}
\end{figure}

  Figure \ref{noise} shows that as the Gaussian noise increases, the accuracy of VQA decreases across all configurations. However, SigLIP + DFT + LoRA consistently outperforms the baseline and LoRA-only models, maintaining a 4-5\% higher accuracy at a noise variance of 0.2.  Figure \ref{blue} illustrates how the BLEU score changes with varying ranks (r) for the DFT + LoRA approach compared to the baseline and LoRA-only configurations. The baseline configuration shows a flat performance, unaffected by rank as it does not incorporate low-rank adaptations. For SigLIP + LoRA, there is a slight improvement as the rank increases, but it plateaus beyond a certain point. The SigLIP + DFT + LoRA configuration exhibits a sharp increase in the BLEU score with higher ranks.

\begin{table*}[!ht]
\caption{The real-time setup uses a camera mounted on the UGV (Husky) (left image). The responses above are from our model and Llava7B, based on the remaining four images, excluding the UGV picture.}
\centering
\renewcommand{\familydefault}{1.2} %
\setlength{\tabcolsep}{4pt} 
\resizebox{\textwidth}{!}{%
\begin{tabular}{|p{16.8cm}|} 
\hline

\vspace{0.1pt}\includegraphics[width=\linewidth]{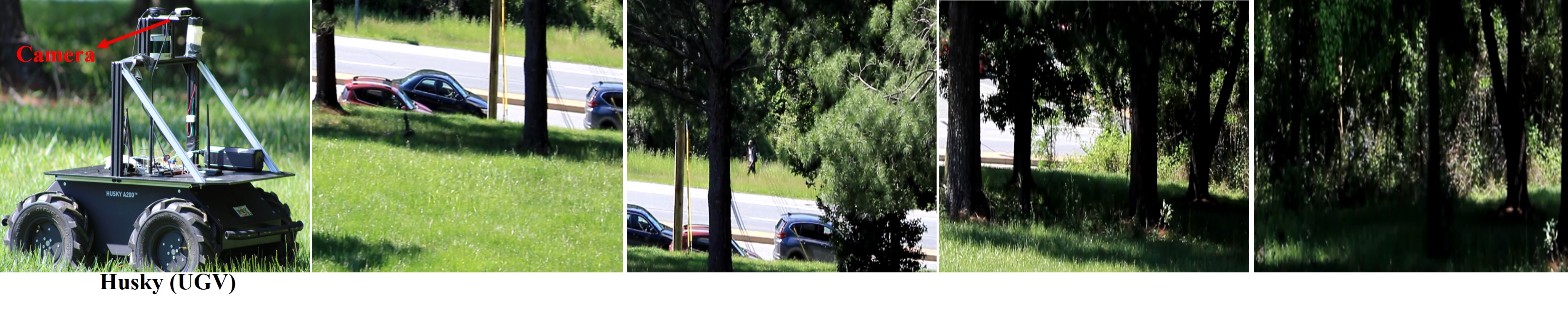} \\ 
\hline

\textbf{Question:} Describe the image in one sentence \textbf{(Same Question for Four Images)} \\
\hline

\textbf{Our model-3B (Image 2):} A \textbf{red car} parks on the side of a road, with a \textbf{black car} and a \textbf{blue car} parked in the background, and a \textbf{yellow pole} in the foreground. \\

\textbf{Llava7B:} The image shows a street scene with cars, trees, and a green grassy area. \\
\hline

\textbf{Our model-3B (Image 3):} A \textbf{person} walks along a tree-lined street, passing a parked car and a parked truck, with a clear blue sky overhead. \\

\textbf{Llava7B:} The image is divided into two halves, showing a car driving down a road surrounded by trees and lawn. \\
\hline

\textbf{Our model-3B (Image 4):} A lush green field with tall trees and a road in the background, creating a serene and tranquil atmosphere. \\

\textbf{Llava7B:} The image shows a street scene with cars, trees, and a green grassy area. \\
\hline

\textbf{Our model-3B (Image 5):} A \textbf{dense forest} with tall trees and a grassy area is bathed in sunlight, creating a serene and tranquil atmosphere. \\

\textbf{Llava7B:} This is an image of a tree-lined street with two pictures side by side. \\
\hline

\end{tabular}%
}

\label{recorded_video_response}
\end{table*}


\begin{table*}[!ht]
\caption{Model Responses to Images Captured from a Camera Feed Mounted on a UGV in a Parking Lot.}
\centering
\renewcommand{\familydefault}{1.2} 
\setlength{\tabcolsep}{4pt} 
\resizebox{\textwidth}{!}{%
\begin{tabular}{|p{16.8cm}|} 
\hline

\vspace{0.1pt}\includegraphics[width=\linewidth]{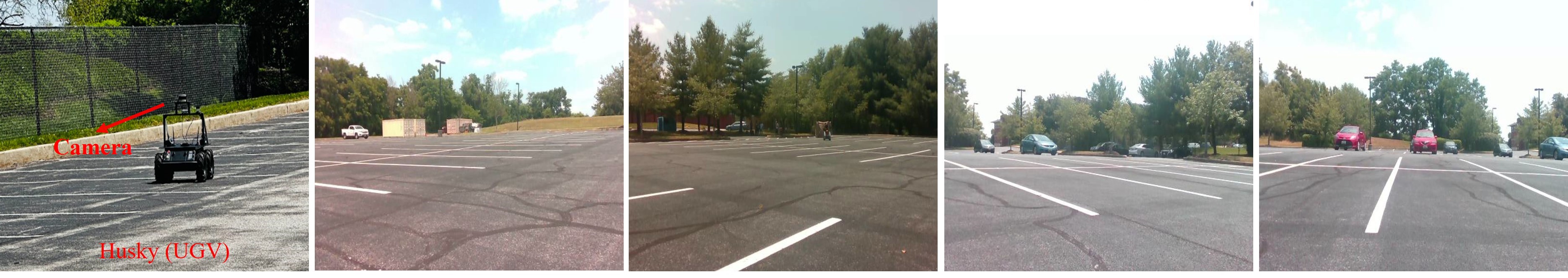} \\ 
\hline

\textbf{Question:} What do you see in this image? Respond in one sentence. \textbf{Same Question for Four Images.} \\
\hline

\textbf{Our model-3B (Image 2):} A large parking lot with a \textbf{white truck} and a small \textbf{white building} in the background. \\

\textbf{Llava7B:} A parking lot with a few cars, trees, and an open sky. \\
\hline

\textbf{Our model-3B (Image 3):} A large parking lot with \textbf{white lines} is surrounded by trees, with a \textbf{few cars} parked in the lot. \\

\textbf{Llava7B:} A mostly empty parking lot with some cars parked. \\
\hline

\textbf{Our model-3B (Image 4):} A parking lot with a \textbf{green car} and a \textbf{silver car}, surrounded by trees and a building. \\

\textbf{Llava7B:} A nearly empty parking lot. \\
\hline

\textbf{Our model-3B (Image 5):} A \textbf{red car} is parked in a parking lot with a white stripe, surrounded by other cars and trees. \\

\textbf{Llava7B:} A parking lot with two vehicles parked next to each other. \\
\hline

\end{tabular}%
}

\label{realtime-response}
\end{table*}

The qualitative results presented in these Tables \ref{recorded_video_response}, \ref{realtime-response}, \ref{online-response}, respectively, demonstrate the effectiveness of our proposed method in the generation of captions and visual question answering (VQA). Through visual examples, we compare the output of our model with the Llava 7B model, highlighting the improvements in contextual accuracy, visual detail preservation, and the ability to generate coherent and contextually appropriate captions.

Table \ref{recorded_video_response} presents the qualitative results of our model on selected frames extracted from a video recorded using a smartphone. The recorded video link and the corresponding model responses are available \textbf{here}. For comparison, we evaluated our model alongside the Llava 7B model asking both models the same question: \textbf{"Describe the image in one sentence."} The first response corresponds to our model and the second to Llava 7B. Our model provides detailed output, including the objects present in the image along with their colors, while Llava 7B fails to detect such details. In practical applications, this difference is significant. For instance, if the robot needs to be instructed to "go to the red car" or "follow the red car," our model's response can be directly used to issue precise commands, unlike Llava 7B. Similar observations can be made across other images, where our model consistently outperforms Llava 7B in providing detailed and actionable outputs.

Next, we present in Table \ref{realtime-response} our model's response to real images captured by a RGB camera.  A RealSense camera was mounted on top of the UGV to capture images in a parking environment. In this scenario, we evaluated the responses of our model and the Llava 7B model using four different images. For each image, we asked the question: "What do you see in this image? Respond in one sentence."
Our proposed model provides more detailed descriptions of the robot's surroundings, offering specific information about objects and their characteristics.  After describing an image, we also asked follow-up questions to gather more information about the surroundings.
For example, questions like "How many cars are present?" or "Do you see any people?" were posed to assess the model's contextual understanding. This level of detail helps our model identify specific objects, making it valuable for real-world robotic tasks.

Table \ref{online-response} presents the responses of our model to a set of images obtained online. To analyze the image content, we asked both models the following question: "Describe the image in a few sentences." The responses of both models are shown, providing a more generalized perspective. Our model generates outputs based solely on the image features, describing exactly what is visible in the image without adding unnecessary details. For example, in the first image, our model describes a bridge that has collapsed onto a ship in the sea or near a harbor. It does not infer the location or provide unrelated details. This is reflected in the response, which focuses strictly on the visible scene. In contrast, the Llava 7B response includes additional details, such as identifying the location as Baltimore and mentioning the height and completion time of the bridge. These details, while interesting, are irrelevant when analyzing real-time images, as observed in Table \ref{realtime-response}, where our model demonstrated its practicality in real-world scenarios by avoiding redundant information. Similarly, for the remaining four images, our model's responses are focused, concise, and relevant, providing a stark contrast to the broader, less applicable information generated by Llava 7B.

\begin{table*}[h!]
\caption{Examples of visual inputs with corresponding user questions and responses from our model and Llava7B, based on online image sources. Each row represents a unique scenario, with key responses highlighted in bold for both models. Our model provides more descriptive and contextually relevant answers. \textbf{Image Source:} \texttt{online}}
\centering
\renewcommand{\familydefault}{1.2} %
\setlength{\tabcolsep}{4pt} 
\resizebox{\textwidth}{!}{%

\begin{tabular}{|p{16.8cm}|} 
\hline

\vspace{0.1pt}\includegraphics[width=\linewidth]{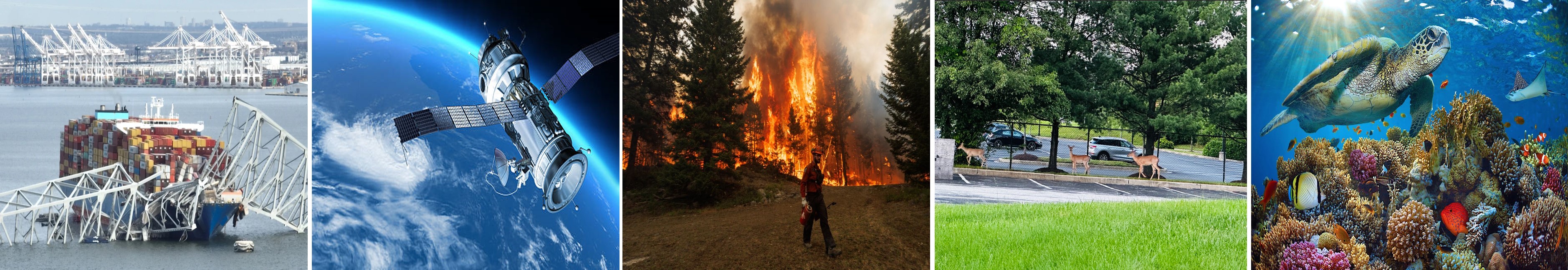} \\ 
\hline

\textbf{Question:} Describe the image in a few sentences. \textbf{Same Question for Five Images.} \\
\hline

\textbf{Our model-3B (Image 1):} A large \textbf{cargo ship}, partially submerged in water, lies on its side with a crane lifting it, while a white crane stands nearby. The scene is set against a backdrop of other cranes and ships in the distance. \\

\textbf{Llava7B:} The image depicts a photograph of the \textbf{Baltimore Harbor Bridge}, also known as the \textbf{Francis Scott Key Bridge}, located in Baltimore, Maryland. The bridge spans across the Chesapeake Bay, connecting downtown Baltimore to Fort McHenry. It is a steel truss bridge completed in \textbf{1973}, standing about 200 feet above the water. \\
\hline

\textbf{Our model-3B (Image 2):} The image features a \textbf{satellite} in the sky, positioned above the Earth. The satellite appears to be a \textbf{weather satellite}, equipped with a weather antenna extending outward. The satellite is flying through the atmosphere, giving the impression of being in space. \\

\textbf{Llava7B:} The image shows an \textbf{artist's representation of a spacecraft}, appearing to be in orbit around a planet. In the foreground, Earth is depicted with blue oceans and clouds, while the background features a clear sky with stars. The satellite, likely powered by \textbf{solar panels}, is positioned at the center, resembling a NASA spacecraft for exploration. \\
\hline

\textbf{Our model-3B (Image 3):} The image depicts a man walking through a \textbf{forest}, with a large fire burning behind him. The \textbf{fire} is intense and consuming trees in its path. The man is carrying a fire hose, likely a \textbf{fire extinguisher}, navigating the area as the fire spreads rapidly, creating a dangerous situation. \\

\textbf{Llava7B:} The image is a realistic photograph of a \textbf{wildfire}, showing flames consuming vegetation. The scene captures the intense fire situation and the environmental impact, set against a night sky with stars in the background. \\
\hline

\textbf{Our model-3B (Image 4):} The image features a group of \textbf{three deer} standing on grass near a parking lot. Positioned close to a fence, one deer stands on the left, another in the middle, and the third on the right side of the scene, with several cars visible in the background. \\

\textbf{Llava7B:} The image shows an outdoor scene where \textbf{two deer} are walking across a paved parking lot. In the background, a chain-link fence, trees, and a building can be seen, suggesting the area is near an urban park or \textbf{recreational space}. \\
\hline

\textbf{Our model-3B (Image 5):} The image depicts a vibrant underwater scene featuring a \textbf{large green sea turtle} swimming among colorful coral and fish. Positioned in the center, the turtle is surrounded by coral in shades of \textbf{red, orange, and yellow}, with sunlight creating a lively atmosphere. \\

\textbf{Llava7B:} The image features an underwater scene showcasing marine life, with a sea turtle swimming close to the viewer. Surrounding the turtle are colorful coral and tropical fish, highlighting the diversity of reef ecosystems. \\
\hline

\end{tabular}%
}

\label{online-response}
\end{table*}

 \section{Conclusion and Future research}
\label{con}

In this research, we proposed an enhanced vision module architecture that integrates frequency-domain features using DFT with low-rank adaptations (LoRA) to improve the performance of vision language models (VLMs). Through extensive experimentation, we evaluated our approach on caption generation and Visual Question Answering (VQA) tasks. The results demonstrate the potential of incorporating frequency domain transformations to enhance the contextual understanding of visual scenes, as evidenced by qualitative results on real-time images captured from a UGV ground robot.
Our approach achieved higher VQA accuracy and evaluation metrics for caption generation compared to baseline models when Gaussian noise levels varied to a variance of 0.2. Incorporating DFT features with LoRA effectively captures both spatial and frequency-domain information. This makes the model more resilient to noise and improves its ability to understand contextual relationships.
Moving forward, our future work will focus on the implementation of the proposed model in low-power edge devices, considering memory capability, power consumption, throughput, and latency. We will also explore additional modalities, such as thermal imaging and lidar, to enhance multimodal understanding and assist decision making, particularly in inaccessible or hazardous environments, such as disaster zones or areas with extreme conditions.

\section{Acknowledgment}
This work is supported by U.S. Army Grant No. W911NF2120076.
\bibliographystyle{unsrt}
\bibliography{bib1}
\end{document}